\documentclass[conference]{IEEEtran}
\IEEEoverridecommandlockouts
\usepackage{cite}
\usepackage{amsmath,amssymb,amsfonts}
\usepackage{algorithmic}
\usepackage{graphicx}
\usepackage{textcomp}
\usepackage{xcolor}
\usepackage[square,numbers]{natbib}
\usepackage{algorithm}
\usepackage{algorithmic}
\usepackage{cleveref}
\usepackage[lofdepth,lotdepth]{subfig}
\usepackage[font=small,labelfont=bf]{caption}
\usepackage{enumitem}
\usepackage[utf8]{inputenc}
\def\BibTeX{{\rm B\kern-.05em{\sc i\kern-.025em b}\kern-.08em
    T\kern-.1667em\lower.7ex\hbox{E}\kern-.125emX}}

\definecolor{darkgreen}{RGB}{30, 160, 80}

\usepackage{fancyhdr,lipsum}
\fancypagestyle{firstpage}{
  \fancyhf{}
  \fancyhead[C]{To appear at the 2020 International Joint Conference on Neural Networks (IJCNN), Glasgow, Scotland, July 2020.}
  \fancyfoot[C]{\thepage}
}

\begin{document}

\title{Is Spiking Secure? A Comparative Study \\on the Security Vulnerabilities of Spiking \\and Deep Neural Networks\\
\vspace{0mm}
}

\author{\IEEEauthorblockN{Alberto Marchisio$^1$, Giorgio Nanfa$^{1,2}$, Faiq Khalid$^1$, Muhammad Abdullah Hanif$^1$,\\
Maurizio Martina$^2$, Muhammad Shafique$^1$}
\vspace*{6pt}
\IEEEauthorblockA{\textit{$^1$Technische Universität Wien, Vienna, Austria}}
\IEEEauthorblockA{\textit{$^2$Politecnico di Torino, Turin, Italy} \vspace*{6pt}\\
Email: \{alberto.marchisio, faiq.khalid, muhammad.hanif, muhammad.shafique\}@tuwien.ac.at}
giorgio.nanfa@studenti.polito.it, maurizio.martina@polito.it
}


\maketitle
\thispagestyle{firstpage}

\begin{abstract}

Spiking Neural Networks (SNNs) claim to present many advantages in terms of biological plausibility and energy efficiency compared to standard Deep Neural Networks (DNNs). Recent works have shown that DNNs are vulnerable to adversarial attacks, i.e., small perturbations added to the input data can lead to targeted or random misclassifications. In this paper, we aim at investigating the key research question: ``Are SNNs secure?'' Towards this, we perform a comparative study of the security vulnerabilities in SNNs and DNNs w.r.t. the adversarial noise. Afterwards, we propose a novel black-box attack methodology, i.e., without the knowledge of the internal structure of the SNN, which employs a greedy heuristic to automatically generate imperceptible and robust adversarial examples (i.e., attack images) for the given SNN. We perform an in-depth evaluation for a Spiking Deep Belief Network (SDBN) and a DNN having the same number of layers and neurons (to obtain a fair comparison), in order to study the efficiency of our methodology and to understand the differences between SNNs and DNNs w.r.t. the adversarial examples. Our work opens new avenues of research towards the robustness of the SNNs, considering their similarities to the human brain's functionality.
\end{abstract}

\vspace*{3pt}

\begin{IEEEkeywords}
Machine Learning, Neural Networks, Spiking Neural Networks, Security, Adversarial Examples, Attack, Vulnerability, Resilience, SNN, DNN, Deep Neural Network.
\end{IEEEkeywords}



\section{Introduction}
Spiking Neural Networks (SNNs), the third generation neural network models~\cite{Maas1997ThirdGenerationSNN}, are rapidly emerging as another design option compared to Deep Neural Networks (DNNs), due to their inherent model structure and properties matching the closest to today's understanding of a brain's functionality. As a result, SNNs have the following key properties. 

\vspace*{3pt}

\begin{itemize}

    \item \textbf{Biologically Plausible}: spiking neurons are very similar to the biological ones because they use discrete spikes to compute and transmit information. For this reason, SNNs are also highly sensitive to the temporal characteristics of the processed data~\cite{Gerstner2002SpikingNeuronModel}\cite{Vreeken2003SNNIntroduction}.
    \item \textbf{Computationally more Powerful than several other NN Models:} a lower number of neurons is required to realize/model the same computational functionality~\cite{Heiberg2013FiringrateSNN}.
    \item \textbf{High Energy Efficiency:} spiking neurons process the information only when a new spike arrives. Therefore, they have relatively lower energy consumption compared to complex DNNs, because the spike events are sparse in time~\cite{Davies2018Loihi}\cite{Merolla2014TrueNorth}\cite{TAVANAEI2019DLSNN}. Such a property makes the SNNs particularly suited for deep learning-based systems where the computations need to be performed at the edge, i.e., in a scenario with limited hardware resources~\cite{Marchisio2019DL4EC}.
    
\end{itemize}

\vspace*{3pt}

SNNs have primarily been used for tasks like real-data classification, biomedical applications, odor recognition, navigation and analysis of an environment, speech and image recognition~\cite{Lisitsa2017ProspectsSNN}\cite{Ponulak2011IntroductionSNN}. 
Recently, the work of~\citeauthor{Fatahi2016evtMNIST}~\cite{Fatahi2016evtMNIST} proposed to convert every pixel of the images into spike trains (i.e., the sequences of spikes) according to its intensity. Since SNNs represent a fundamental step towards the idea of creating an architecture as similar as possible to the current understanding of the structure of a human brain, \textit{it is fundamental to study their security vulnerability w.r.t. adversarial attacks. In this paper, we demonstrate that indeed, even a small adversarial perturbation of the input images can modify the spike propagation and increase the probability of the SNN misprediction (i.e., the image is misclassified)}.

\vspace*{6pt}

\textbf{Adversarial Attacks on DNNs:} In recent years, many methods to generate adversarial attacks for DNNs and their respective defense techniques have been proposed~\cite{Goodfellow2015AdversarialExamples}\cite{Kurakin2017AdversarialPhysicalworld}\cite{Madry2018DLResistantAdversarial}. A minimal and imperceptible modification of the input data can cause a classifier misprediction, which can potentially produce a wrong output with high probability. This scenario may lead to serious consequences in safety-critical applications (e.g., automotive, medical, UAVs and banking) where even a single misclassification can have catastrophic consequences~\cite{Zhang2019RobustML}.

\vspace*{6pt}

In the image recognition field, having a wide variety of possible real-world input images~\cite{Kurakin2017AdversarialPhysicalworld}, with diverse pixel intensity patterns, the classifier cannot recognize if the source of the misclassification is the attacker or other factors~\cite{Shafahi2018AdversarialInevitable}. Given an input image $x$, the goal of an adversarial attack \mbox{$x^*=x+\delta$} is to apply a small perturbation $ \delta $ such that the predicted class $C(x)$ is different from the target one $C(x^*)$, i.e., the class in which the attacker wants to classify the example. Inputs can also be misclassified without specifying the target class. This is the case of untargeted attacks, where the target class is not defined a-priori by the attacker. Targeted attacks can be more difficult to apply than the untargeted ones, but they can be relatively more effective in several cases~\cite{Zhang2018AdversarialOpportunitiesChallenges}. Another important classification of adversarial attacks is based on the knowledge of the network under attack, as discussed below.

\vspace*{3pt}

\begin{itemize}
    \item \textit{White-box attack:} an attacker has the complete access and knowledge of the architecture, the network parameters, the training data and the existence of a possible defense.
    \item \textit{Black-box attack:} an attacker does not know the architecture, the network parameters, the training data and a possible defense, but it can only access to the input and output of the network (which is treated as a black-box), and may be the testing dataset~\cite{Papernot2017PracticalBlackBox}. 
\end{itemize}

\vspace*{6pt}

\textbf{Our Approach towards Adversarial Attacks on SNNs: }\textit{In this paper, we aim at generating, for the first time, imperceptible and robust adversarial examples for SNNs under the black-box settings.}~\citeauthor{Bagheri2018AdvTrainingSNN}~\cite{Bagheri2018AdvTrainingSNN} studied the vulnerabilities of SNNs under white-box assumptions, while we consider a black-box scenario, which makes the attacker stronger under a wide range of real-world scenarios. For the evaluation, we apply these attacks to a Spiking Deep Belief Network (SDBN) and a DNN having the same number of layers and neurons, to obtain a fair comparison. As per our knowledge\footnote{A previous version of this work is available in~\cite{Marchisio2019SNNUnderAttack}.}, this kind of black-box attack was previously applied \textbf{only} to a DNN model~\cite{Luo2018ImperceptibleRobust}. This method is efficient for DNNs because it is able to generate adversarial noise which is imperceptible to the human eye.

\vspace*{6pt}

\begin{figure}[t]
\vspace*{0mm}
\centering
\includegraphics[width=.95\linewidth]{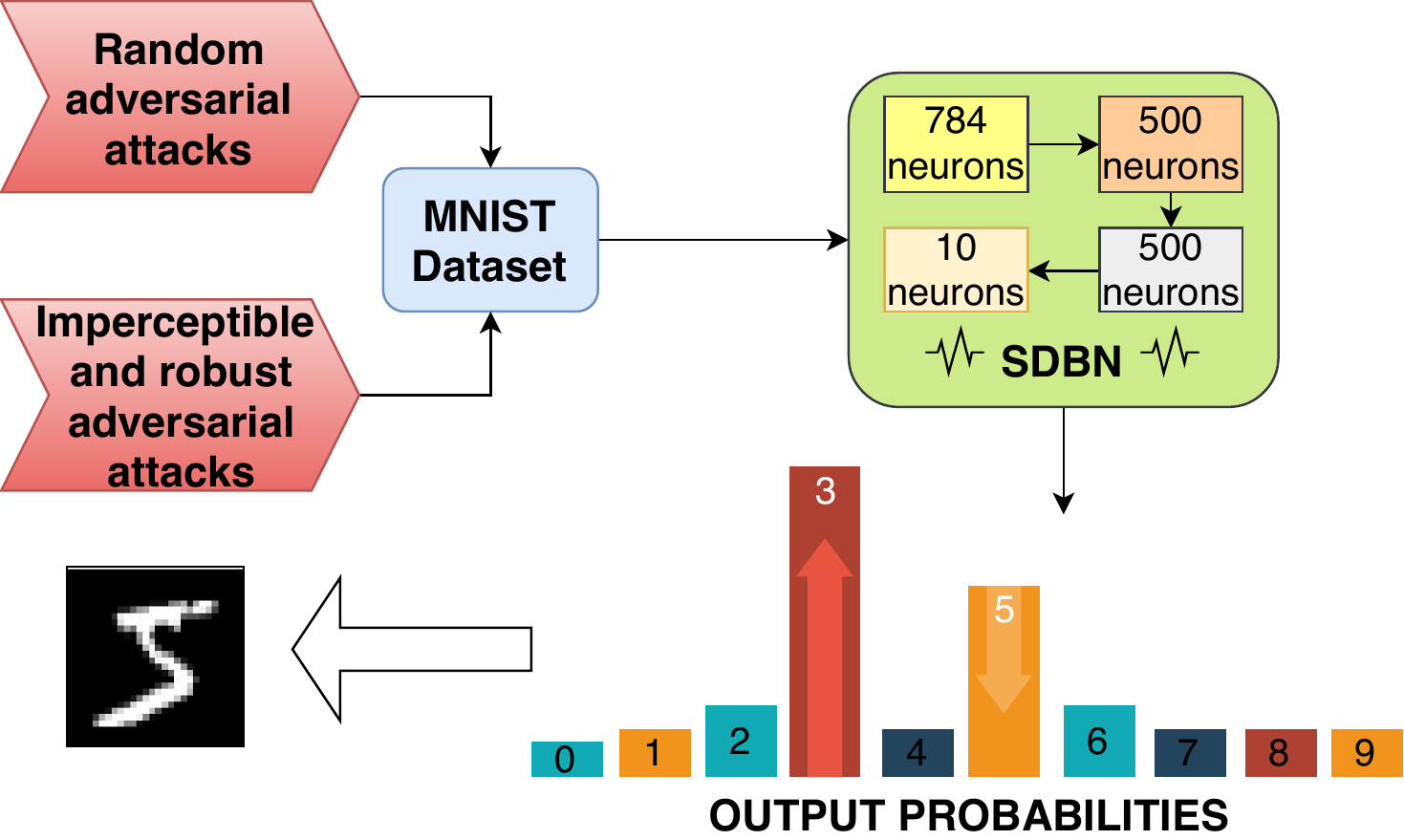}
\caption{Overview of our proposed approach.}
\label{fig:intro_fig}
\end{figure}

As shown in Figure~\ref{fig:intro_fig}, we investigate the vulnerability of SDBNs to random noise and adversarial attacks, aiming at identifying the similarities / differences w.r.t. DNNs. Our experiments show that, when applying a random noise to a given SDBN, its classification accuracy decreases, by increasing the noise magnitude. Moreover, when applying our attack to SDBNs, we observe that, in contrast to the case of DNNs, the output probabilities follow a different behavior, i.e., while the adversarial image remains imperceptible, the misclassification is not always guaranteed.

\vspace*{6pt}

\textbf{In short, we make the following Novel Contributions:}

\vspace*{3pt}

\begin{enumerate}
    \item We analyze the variation in the accuracy of a Spiking Deep Belief Network (SDBN) when a random noise is added to the input images. (\textbf{Section~\ref{sec:randomnoise}})
    \item We evaluate the improved generalization capabilities of the SDBN when adding a random noise to the training images. (\textbf{Section~\ref{subsec:noise_window}})
    \item We develop a methodology to automatically create imperceptible adversarial examples for SNNs. (\textbf{Section~\ref{sec:attackdesign}})
    \item We apply our methodology to an SDBN \textit{(it is the first attack of this type applied to SDBNs)} and a DNN for generating adversarial examples, and evaluate their imperceptibility and robustness. (\textbf{Section~\ref{sec:results}})
\end{enumerate}

\vspace*{6pt}

Before proceeding to the technical sections, in \textbf{Section~\ref{sec:related}}, we briefly discuss the background and the related work, focusing on SDBNs and adversarial attacks on DNNs.

\vspace*{6pt}

\section{Background and Related Work}
\label{sec:related}

\vspace*{3pt}

\subsection{Spiking Deep Belief Networks}
Deep Belief Networks (DBNs)~\cite{Bengio2007GreedyTraining} are multi-layer networks that are widely used for classification problems and have been implemented in many areas such as visual processing, audio processing, images and text recognition~\cite{Bengio2007GreedyTraining}. DBNs are implemented by stacking pre-trained Restricted Boltzmann Machines (RBMs), energy-based models consisting in two layers of neurons, one hidden and one visible, symmetrically and fully connected, i.e., without connections between the neurons inside the same layer (this is the main difference w.r.t. the standard Boltzmann machines). RBMs are typically trained with unsupervised learning, to extract the information saved in the hidden units, and then a supervised training is performed to train a classifier based on these features~\cite{Hinton2006ReducingDimensionalityDNN}. 

\vspace*{6pt}

\textit{Spiking DBNs (SDBNs) improve the energy efficiency and computation speed, as compared to DBNs}. Such a behavior has already been observed by~\citeauthor{OConnor2013RTSNN}~\cite{OConnor2013RTSNN}. That work proposed a DBN model composed of 4 RBMs of 784-500-500-10 neurons, respectively. It has been trained offline and transformed in an event-based domain to increase the processing efficiency and the computational power. The RBMs are trained with the Persistent Contrastive Divergence (CD) algorithm, an unsupervised learning rule using Gibbs sampling, a Markov-Chain Monte-Carlo algorithm, with optimizations for fast weights, selectivity and sparsity~\cite{Goh2010BiasingRBM}\cite{Merino2018WeightedContrastiveDivergence}\cite{Tieleman2009FastWeights}. Once every RBM is trained, the information is stored in the hidden units to use it as an input for the visible units of the following layer. Afterwards, a supervised learning algorithm~\cite{Hinton2006FastLearningDBN}, based on the features coming from the unsupervised training, is performed. The RBMs of this model use the \textit{Siegert} function~\cite{Siegert1951FirstPassageProbability} in their neurons. It allows to have a good approximation of firing rate of Leaky Integrate and Fire (LIF) neurons~\cite{Gerstner2002SpikingNeuronModel}, used for CD training. Hence, the neurons of an SDBN generate Poisson spike trains, according to the \textit{Siegert} formula.

\vspace*{6pt}

This represents a great advantage in terms of power consumption and speed, as compared to the classical DBNs, which are based on a discrete-time model~\cite{OConnor2013RTSNN}. \textit{Since there has been no prior work on studying the security vulnerabilities of SNNs / SDBNs, we aim at investigating these aspects in a black-box setting, which is important for their real-world applications in security/safety-critical systems}.

\vspace*{6pt}

\subsection{Adversarial Attacks for DNNs}

The robustness and self-healing properties of DNNs have been thoroughly investigated in the recent researches~\cite{Shafique2020RobustML}. As demonstrated for the first time by~\citeauthor{Szegedy2014IntriguingPropNN}~\cite{Szegedy2014IntriguingPropNN}, adversarial attacks can misclassify an image by changing its pixels with small perturbations. \citeauthor{Kurakin2017AdversarialPhysicalworld}~\cite{Kurakin2017AdversarialPhysicalworld} defined adversarial examples as \textit{a sample of input data which has been modified very slightly in a way that is intended to cause a machine learning classifier to misclassify it}. \citeauthor{Luo2018ImperceptibleRobust}~\cite{Luo2018ImperceptibleRobust} proposed a method to generate attacks by maximizing their noise tolerance and taking into account the human perceptual system in their distance metric. A similar attack is able to mislead even more complex DNNs, like Capsule Networks~\cite{Marchisio2019CapsAttacks}, which are notoriously more robust against adversarial attacks. This methodology has strongly inspired our algorithm. The human eyes are more sensitive to the modifications of the pixels in low variance areas. Hence, to maintain the imperceptibility as much as possible, the modification of pixels in only the high variance areas is preferable.

\vspace*{6pt}

Moreover, a robust attack aims at increasing \textit{its ability to stay misclassified to the target class after the transformations due to the physical world}. For example, considering a crafted sample, after an image compression or a resizing, its output probabilities can change according to the types of the applied transformations. Therefore, the attack can be ineffective if it is not robust enough to those variations.

\vspace*{6pt}

Motivated by the above-discussed considerations, \textit{we propose an algorithm to automatically generate imperceptible and robust adversarial examples for SNNs, and study their differences w.r.t. the adversarial examples generated for DNNs using the same technique}.

\vspace*{6pt}

\section{Analysis: Applying Random Noise to SDBNs}
\label{sec:randomnoise}

\vspace*{3pt}

\subsection{Experimental Setup}
For a case study, we consider an SDBN~\cite{OConnor2013RTSNN} composed of four fully-connected layers of 784-500-500-10 neurons, respectively. We implement this SDBN in Matlab, for analyzing the MNIST database, a collection of 28$\times$28 gray scale images of handwritten digits, divided into 60,000 training images and 10,000 test images. Each pixel intensity is encoded as a value between 0 and 255. To maximize the spike firing, the input data are scaled to the range [0,0.2], before converting them into spikes. In our experiments, the pixel intensities are represented as the probability that a spike occurs.

\vspace*{6pt}

\subsection{Understanding the Impact of Random Noise Addition to Inputs on the Accuracy of an SDBN}
\label{subsec:random_noise}

We test the accuracy of the SDBN for different noise magnitudes, applied to three different combinations of images:

\vspace*{3pt}

\begin{itemize}
    \item to all the training images only.
    \item to all the test images only.
    \item to both the training and test images.
\end{itemize}

\vspace*{6pt}

To test the vulnerability of the SDBN, we apply two different types of noises: \textit{normally-distributed} and \textit{uniformly-distributed} random noise.

\vspace*{6pt}

\begin{figure}[h]
\vspace*{5pt}
\centering
\includegraphics[width=\linewidth]{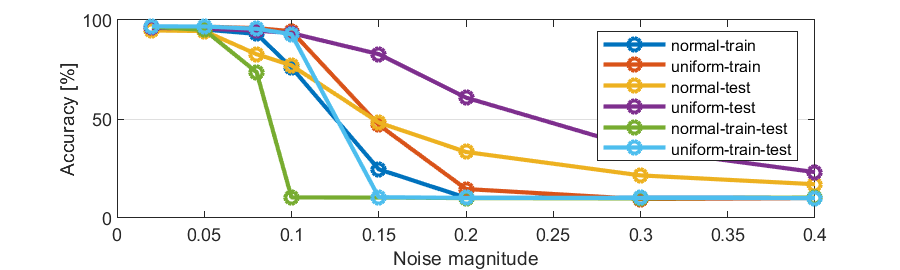}
\caption{Normal and uniform random noise applied to all the pixels of the MNIST dataset.}
\label{fig:noise_all}
\end{figure}

The results of our experiments are shown in \Cref{tab:my_table} and Figure~\ref{fig:noise_all}. The initial ``clean-case'' accuracy, obtained without applying noise, is $96.2\%$. When the noise is applied to the test images, the accuracy of the SDBN decreases accordingly with an increase in the noise magnitude, more evidently in the case of the normally-distributed random noise. This behavior is due to the fact that the standard normal distribution contains a wider range of values, compared to the uniform distribution. For both noise distributions, the accuracy decreases more when the noise magnitude applied is around 0.15 (see the red-colored values in \Cref{tab:my_table}).

\begin{table}[h]
\vspace*{2mm}
    \caption{Evaluation of the SDBN accuracy applying two different types of random noise with different values of noise magnitude. The red and blue values are helping the reader to identify the accuracy results that are discussed in the text. (ACC stands for Accuracy, TR+TST stands for Training and Test Datasets)}
    \label{tab:my_table}
    \centering
\resizebox{\linewidth}{!}{%
\begin{tabular}{|c|c|c|c|c|c|c|}
\hline
       \textbf{ACC} &      \textbf{TRAIN} &       \textbf{TEST} & \textbf{TR+TST} &      \textbf{TRAIN} &       \textbf{TEST} & \textbf{TR+TST} \\
\hline
     $\delta$ &      \multicolumn{ 3}{|c|}{NORMALLY} &    \multicolumn{ 3}{|c|}{ UNIFORMLY} \\
\hline
      0.02 &      \color{blue}\textbf{96.65}\color{black} &      94.73 &      96.54 &       \color{blue}\textbf{96.8}\color{black} &      96.02 &      96.81 \\
\hline
      0.05 &      95.19 &      94.42 &      94.99 &       96.7 &      95.64 &      96.72 \\
\hline
      0.08 &      92.99 &      82.73 &      73.64 &      95.89 &      94.64 &      95.56 \\
\hline
       0.1 &      76.01 &      77.07 &      10.39 &      94.34 &      93.36 &       92.8 \\
\hline
      0.15 &      24.61 &      \color{red}\textbf{48.23}\color{black} &      10.32 &      47.03 &      \color{red}\textbf{82.76}\color{black} &      10.51 \\
\hline
       0.2 &      10.26 &      33.34 &      10.05 &      14.64 &      60.79 &      10.16 \\
\hline
       0.3 &      10.31 &      21.52 &       9.88 &       9.59 &       34.9 &      10.16 \\
\hline
       0.4 &      10.27 &      17.05 &      10.34 &       9.98 &      23.16 &      10.03 \\
\hline
\end{tabular}  

}
    \vspace*{2mm}
\end{table}

\vspace*{6pt}

When the noise is applied to the training images, the accuracy of the SDBN does not decrease as much as in the previous case, as long as the noise magnitude ($\delta$) is lower than 0.1. On the contrary, for $\delta=0.02$, the accuracy increases (see the blue-colored values in \Cref{tab:my_table})  w.r.t. the baseline  (i.e., without noise). Indeed, adding noise in training samples improves the generalization capabilities of the neural network. Hence, its capability to correctly classify new unseen samples also increases. This observation, as was analyzed in several other scenarios for Deep Neural Networks with back-propagation training~\cite{Holmstrom1992AdditiveNoiseBackprop}, is also valid for our SDBN model. However, if the noise is equal to or greater than 0.1, the accuracy drops significantly. This behavior means that the SDBN is unable to learn input features due to the inserted noise, thus it is unable to correctly classify the inputs.

\vspace*{6pt}

When the noise is applied to both the training and test images, we notice that the behavior observed for the case of noise applied to only the training images is accentuated. For low noise magnitudes (mostly in the uniform noise case), the accuracy is similar or higher than the baseline. For noise magnitudes greater than 0.1 (more precisely, 0.08 for the case of normal noise applied), the accuracy decreases more sharply than in the case of noise applied only to the training images. Such a value of noise magnitude represents a threshold of tolerable noise for the SDBN. Hence, when the noise is too high, the network cannot classify well.

\vspace*{3pt}

\subsection{Applying Noise to a Restricted Window of Pixels}
\label{subsec:noise_window}

\begin{figure}[h]
\vspace*{5pt}
\includegraphics[scale=0.6]{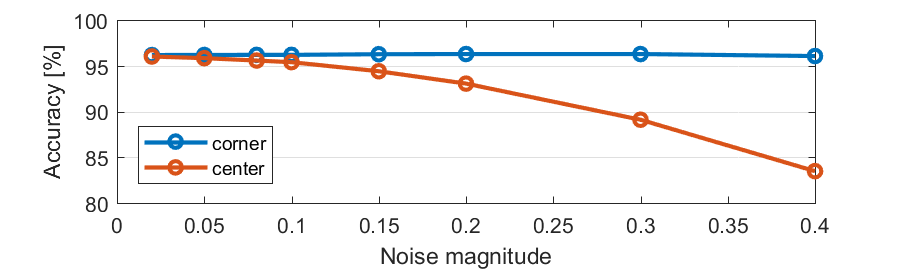}
\caption{Normal random noise applied to some pixels of the MNIST test images.}
\label{fig:noise_rec}
\vspace*{3pt}
\end{figure}

\begin{figure}[h]
\vspace*{0mm}
\centering
\begin{minipage}[t]{.30\linewidth}
\subfloat[]{
\includegraphics[width=.9\linewidth]{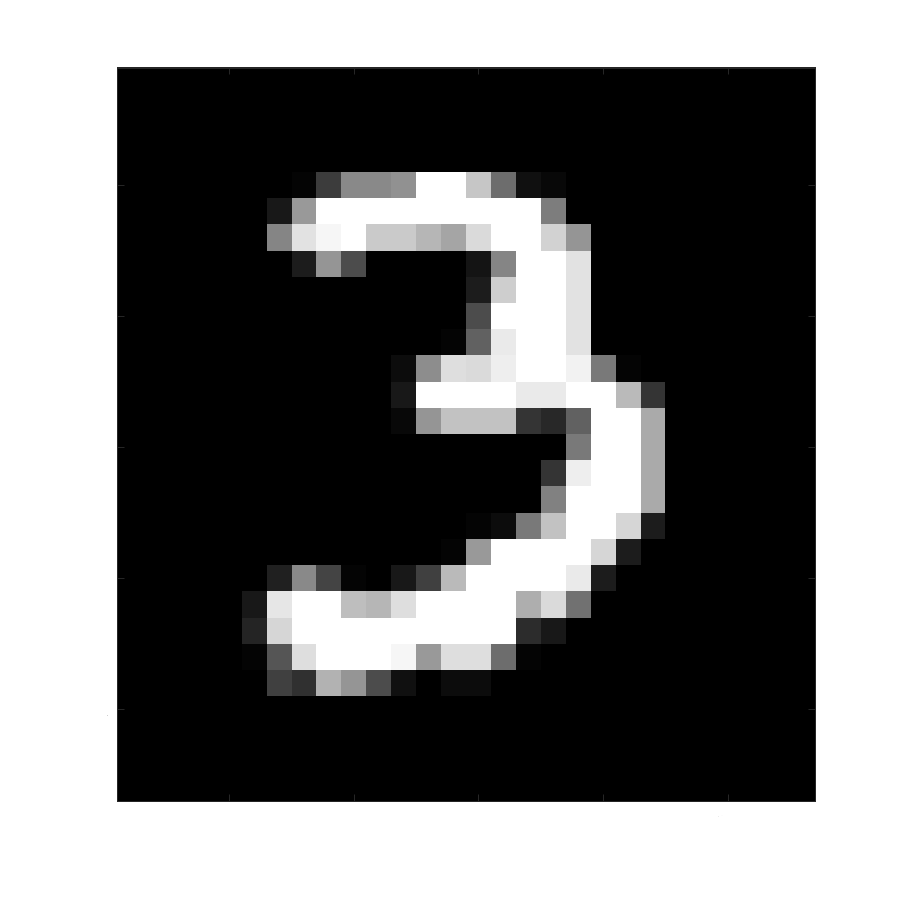}
\vspace*{-5mm}
\label{fig:3nonoise}}
\end{minipage}
\hfill
\begin{minipage}[t]{.30\linewidth}
\subfloat[]{
\includegraphics[width=.9\linewidth]{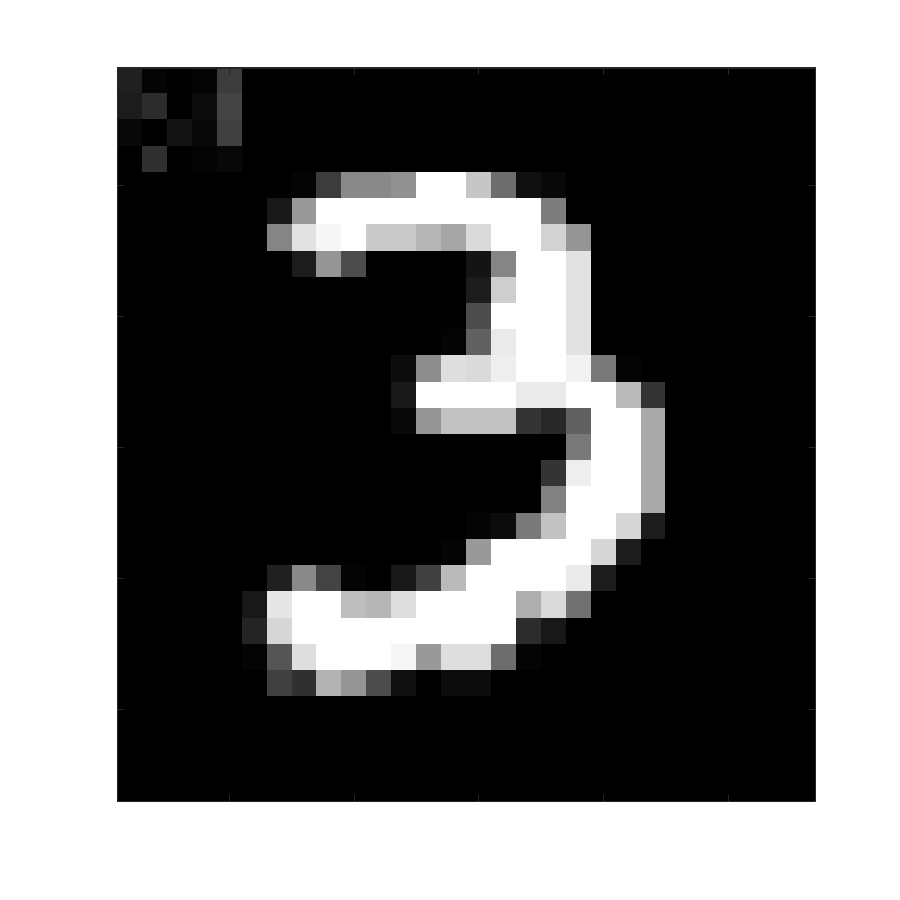}
\vspace*{-5mm}
\label{fig:3noisecorner}}
\end{minipage}
\hfill
\begin{minipage}[t]{.30\linewidth}
\subfloat[]{
\includegraphics[width=.9\linewidth]{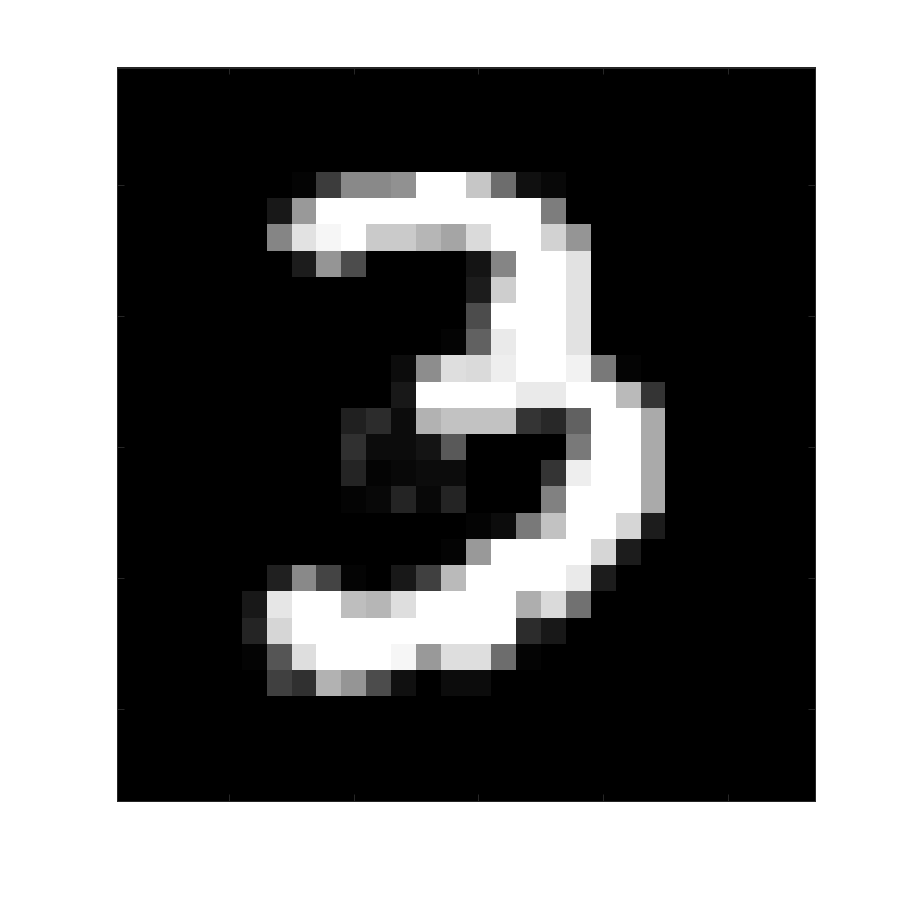}
\vspace*{-5mm}
\label{fig:3noisecenter}}
\end{minipage}
\vspace*{0mm}
\caption{Comparison between images with normally distributed random noise (with magnitude 0.3) applied to the corner and to the left center of the image. (a) Without noise. (b) Noise applied to the top-left corner. (c) Noise applied to the center.}
\label{fig:noise_comparison}
\end{figure}

In this analysis, we add a normally distributed random noise to a restricted window of pixels of the test images. Considering a rectangle of 4$\times$5 pixels, we analyze two scenarios:

\vspace*{3pt}

\begin{itemize}
    \item The noise is applied to 20 pixels at the top-left corner of the image. The variation of the accuracy is represented by the blue-colored line of Figure~\ref{fig:noise_rec}. As expected, the accuracy remains almost constant, because the noise affects irrelevant pixels. The resulting image, when the noise is equal to 0.3, is shown in Figure~\ref{fig:3noisecorner}.
    \item The noise is applied to 20 pixels in the middle of the image, with coordinates $(x,y) = ([14\ 17], [10\ 14])$. The accuracy descreases more significantly (orange-colored line of Figure~\ref{fig:noise_rec}), as compared to the previous case, because some white pixels representing the handwritten digits (and therefore the important ones for the classification) are affected by the noise. The resulting image, when the noise is equal to 0.3, is shown in Figure~\ref{fig:3noisecenter}. This analysis shows that the location of noise insertion impacts the accuracy, thereby unleashing a potential vulnerability of SNNs that can be exploited by the adversarial attacks.
\end{itemize}

\subsection{Key Observations from our Analyses}
\label{subsed:key_obs}

From the analyses performed in the above Sections~\ref{subsec:random_noise} and~\ref{subsec:noise_window}, we derive the following key observations that can be exploited by an adversarial example generation methodology.

\vspace*{3pt}

\begin{itemize}
    \item The normal noise is more powerful than the uniform counterpart, since the accuracy decreases more sharply.
    \item For a low noise magnitude applied to the training images, we notice a small accuracy improvement, due to the improved generalization capability of SDBNs.
    \item When applying the noise to a restricted window of pixels, the perturbation is more effective if the window is in the center of the image (or generally speaking, in the input regions belonging to the  features that are key for the correct classification), as compared to the corner. This is due to the fact that the noise is applied to the pixels which play an important role for accurate feature learning and consequently for the correct classification.
\end{itemize}

\vspace*{6pt}

\section{Our Novel Methodology to Generate Imperceptible and Robust Adversarial Examples}
\label{sec:attackdesign}
\textit{Similar to the case of DNNs, the scope of a good attack on SNNs is also to generate adversarial images, which are difficult to be detected by human eyes (i.e., imperceptible) and resistant to physical transformations (i.e., robust)}. Therefore, for better understanding, we first discuss these two aspects.

\vspace*{6pt}

\subsection{Imperceptibility of Adversarial Examples}

Creating an imperceptible example means to add perturbations to the pixels, while making sure that humans do not notice them. We consider an area A=N$\cdot$N of pixels $x$, and we compute the standard deviation (SD) of a pixel $x_{i,j}$ as in \Cref{eq:SD}.

\vspace*{0mm}
\begin{equation}
    SD(x_{i,j})=\sqrt{\frac{ \sum\limits_{k=1}^N \sum\limits_{l=1}^N (x_{k,l}-\mu)^2-(x_{i,j}-\mu)^2}{N \cdot N}}
    \label{eq:SD}
\end{equation}

\vspace*{6pt}

Here, $\mu$ is the average value of pixels belonging to the N$\cdot$N area. If a pixel has a high standard deviation, it means that a perturbation added to this pixel is more likely to be hardly detected by the human eye, compared to a pixel with a low standard deviation. The sum of all the perturbations $\delta$ added to the pixels of the area A allows to compute the distance ($D(X^*,X)$) between the adversarial example $X^*$ and the original one $X$. Its formula is shown in \Cref{eq:D}.

\vspace*{0mm}
\begin{equation}
    D(X^*,X)=\sum\limits_{i=1}^{N} \sum\limits_{j=1}^{N}\frac{\delta_{i,j}}{SD(x_{i,j})}
    \label{eq:D}
\end{equation} 

\vspace*{3pt}

Such value can be used to monitor the imperceptibility. Indeed, the distance $D(X^*,X)$ indicates how much perturbation is added to the pixels in the area A. Hence, the maximum perturbation tolerated by the human eye can be associated to a certain value of the distance, $D_{MAX}$. The value of $D_{MAX}$ can vary among different datasets or images, because it depends on the resolution and the contrast between neighboring pixels.

\vspace*{6pt}

\subsection{Robustness of adversarial examples}

Many adversarial attack methods used to maximize the probability of target class to ease the classifier misclassification of the image. The main problem of these methods is that they do not account for the relative difference between the class probabilities, i.e., the gap, defined in \Cref{eq:gap}.

\begin{equation}
\resizebox{.85\linewidth}{!}{%
   $Gap(X^*)=P(target\ class)-max\{P(other\ classes)\}$}
   \label{eq:gap}
\end{equation}

\vspace*{3pt}

Therefore, after an image transformation, a minimal modification of the probabilities can make the attack ineffective. To improve the robustness, it is desirable to increase the difference between the probability of the target class and the highest probability of the other classes, i.e., to maximize the gap function.

\vspace*{6pt}

\subsection{How to Automatically Generate Attacks for SNNs?}
\label{subsec:methodology}

Obtaining both the imperceptibility and robustness at the same time is complicated. Typically, a robust attack would require perceptible changes of the input, while an imperceptible attack does not change the classification much. We designed a heuristic algorithm to automatically generate imperceptible yet robust adversarial examples for SNNs. Our technique is also applicable to DNNs, as we will demonstrate in the result section. Note that, leveraging the same methodology to generate adversarial examples for both SNNs and DNNs enables a fair comparison. Our algorithm is based on the black-box model, i.e., the attacks are performed on some pixels of the image, without having insights of the network.
Given the maximum allowed distance $D_{MAX}$ such that human eyes cannot detect perturbations, the problem can be expressed as in \Cref{eq:argmaxgap}.

\vspace*{0mm}
\begin{equation}
    \arg\max_{X^*}Gap(X^*) \,\ |\ \, D(X^*,X) \le D_{MAX}
    \label{eq:argmaxgap}
\end{equation} 

\vspace*{3pt}

In summary, \textit{the purpose of our iterative algorithm is to perturb a set of pixels, to maximize the gap function, thus making the attack robust, while keeping the distance between the samples below the desired threshold, in order to remain imperceptible}.

\vspace*{6pt}

Based on the key observations of our analysis in Section~\ref{subsed:key_obs}, our iterative methodology (see Algorithm~\ref{Greedy algorithm}) perturbs only a window of pixels of the image. We choose a certain value N, which corresponds to an area of N$\cdot$N pixels, performing the attack on a subset M of pixels.

\begin{figure}[h]
\begin{algorithm}[H]
\caption{\textbf{: Methodology for Generating Adversarial Examples for SNNs and DNNs}}\label{Greedy algorithm}
\begin{small}
\begin{algorithmic}
 \STATE Given: network (SNN or DNN), original sample X, maximum human perceptual distance $D_{max}$, noise magnitude $\delta$, area of A pixels, target class, M
\WHILE{$D(X^*,X)<D_{MAX}$}
\STATE -Compute \textit{Standard Deviation SD} for every pixel of A
\STATE -Compute $Gap(X^*)$, $Gap^-(X^*)$, $Gap^+(X^*)$
\IF{$Gap(X^*)^- > Gap(X^*)^+$}
\STATE $VariationPriority(x_{i,j})=$\\$[Gap^-(X^*) - Gap(X^*)] \cdot SD(x_{i,j})$
\ELSE
\STATE $VariationPriority(x_{i,j})=$\\$[Gap^+(X^*) - Gap(X^*)] \cdot SD(x_{i,j})$
\ENDIF
\STATE -Sort in descending order $VariationPriority$
\STATE -Select M pixels with highest $VariationPriority$
\IF{$Gap(X^*)^- > Gap(X^*)^+$}
\STATE Subtract noise with magnitude $\delta$ from the pixel 
\ELSE
\STATE Add noise with magnitude $\delta$ to the pixel 
\ENDIF
\STATE -Compute $D(X^*,X)$
\STATE -Update the original example with the adversarial one
\ENDWHILE
\end{algorithmic}
\end{small}
\end{algorithm}
\end{figure}

\begin{figure*}[h]
\vspace*{0mm}
\centering
\includegraphics[width=.7\linewidth]{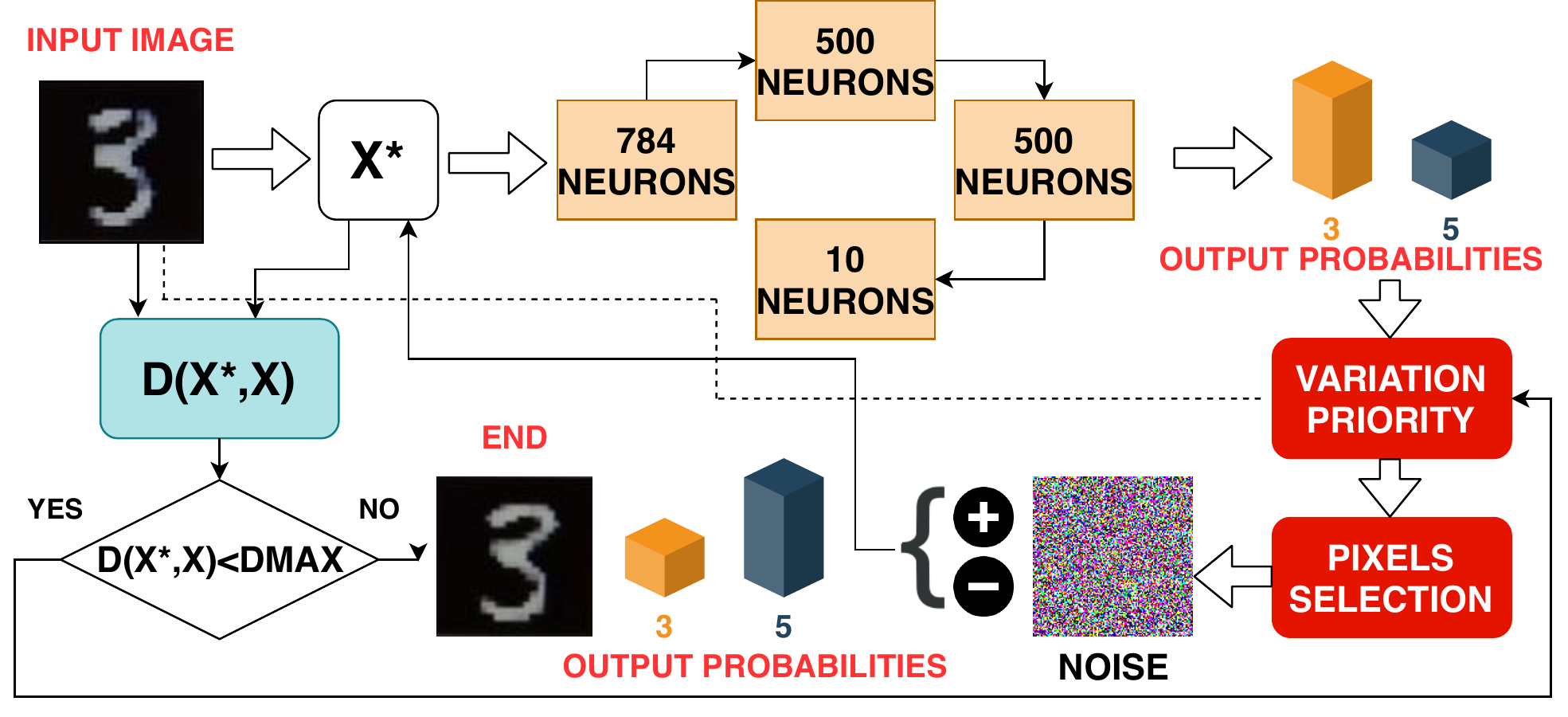}
\vspace*{0mm}
\caption{Our methodology for generating adversarial examples, illustrating with the example of the considered networks.}
\vspace*{0mm}
\label{fig:attack}
\vspace*{2mm}
\end{figure*}

After computing the standard deviation for the selected N$\cdot$N pixels, we compute the gap function, i.e., the difference between the probability of the target class and the highest probability between the other classes. Then, the algorithm decides whether to apply a positive or a negative noise to the pixels. Therefore, we compute two parameters for each pixel, $Gap^+(X^*)$ and $Gap^-(X^*)$. $Gap^+(X^*)$ is the value of the gap function computed by adding a perturbation unit to the single pixel, while $Gap^-(X^*)$ is its counterpart, computed subtracting a perturbation unit. According to the difference between these values and the gap function, and considering also the standard deviation, we compute the variation priority, a function that indicates the effectiveness of the pixel perturbation. For example, if $Gap^-(X^*)$ is greater than $Gap^+(X^*)$, it means that, for the pixel under consideration, subtracting the noise will be more effective than adding it to the pixel, since the difference between $P(target\ class)$ and $max[P(other\ classes)]$ will increase more. Once computed the vector \textit{VariationPriority}, its values are sorted, and the highest M values are perturbed. Note, according to the previous considerations, the noise is added to, or subtracted from, the selected M pixels depending on the highest value between $Gap^+(X^*)$ and $Gap^-(X^*)$. The algorithm starts the next iteration by replacing the original input image with the created adversarial one. The iterations terminate when the distance between original and adversarial examples overcomes the maximum perceptual distance. Figure~\ref{fig:attack} summarizes the operational flow of our methodology, applied to the SDBN, for generating adversarial examples.

\vspace*{6pt}

\section{Evaluating our Attack Methodology on SDBNs and DNNs}
\label{sec:results}

\vspace*{3pt}

\subsection{Experimental Setup}
Using the methodology of Section~\ref{subsec:methodology}, we attack two different networks: the same SDBN as the one analyzed in Section~\ref{sec:randomnoise} and a DNN. To achieve a fair comparison, we design the DNN for our experiments having the same set of parameters as the SDBN, i.e., composed of four fully-connected layers of 784-500-500-10 neurons, respectively. The DNN is trained with the scaled conjugate gradient backpropagation algorithm~\cite{Mller1993ScaledConjugateGradient}, and after training, its achieved classification accuracy on the MNIST dataset is $97.13\%$.

\vspace*{6pt}

\begin{figure}[h]
\centering
\vspace*{6pt}
\subfloat[]{
\includegraphics[scale=0.45]{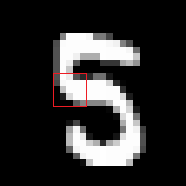}
\label{fig:5_smallwindow}}
\hspace{14mm}
\subfloat[]{
\includegraphics[scale=0.45]{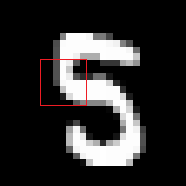}
\label{fig:5_bigwindow}}
\vspace*{0mm}
\caption{Selected area of pixels to attack}
\label{fig:attzone}
\vspace*{0mm}
\end{figure}

For discussion, we start with a test sample, labeled as ``five'' (see \Cref{fig:attzone}). It is classified correctly by both networks, but with different output probabilities. We use a value of $\delta$ equal to the $10\%$ of the pixel intensity scale range and a $D_{MAX}$ equal to 22 to compare the attacks. We distinguish two cases, having different search window sizes:

\vspace*{3pt}

\begin{enumerate}[label=(\Roman*)]
    \item Figure~\ref{fig:5_smallwindow}: N=5 and M=10. Based on the analysis in Section~\ref{sec:randomnoise}, we define the search window in a central area of the image, as shown by the red square, which is affected by high variation.
    \item Figure~\ref{fig:5_bigwindow}: N=7 and M=10. It can be interesting to observe the difference w.r.t. the case I: in this situation we perturb the same amount M of pixels, selected from a search window which contains 24 more pixels.
\end{enumerate}
 
\vspace*{6pt}

\subsection{DNN Under Attack}
\begin{figure}[h]
\vspace*{15pt}
\centering
\includegraphics[width=\linewidth]{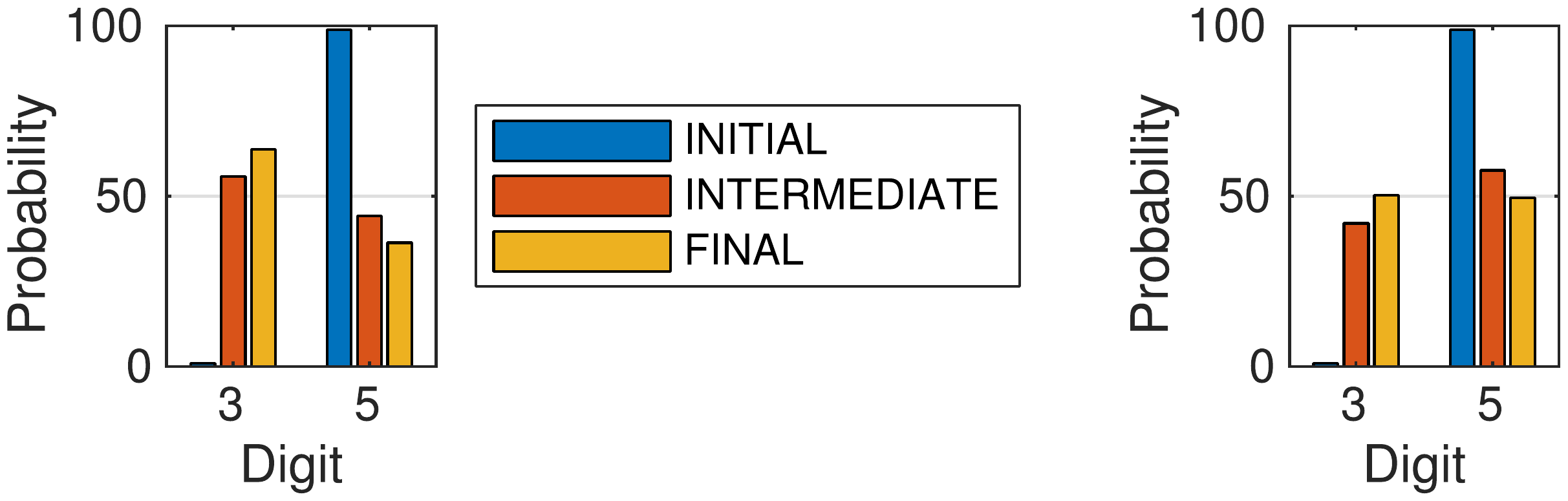}
\label{fig:dnnnoise25}
\\
\begin{footnotesize}
\hspace{14 mm} (a) \hspace{59 mm} (b) \hfill
\end{footnotesize}
\vspace*{0mm}
\caption{Output probabilities (\% format) of the DNN. (a) Attack using the search window of case I. (b) Attack using the search window of case II.}
\vspace*{0mm}
\label{fig:dnnnoise_both}
\vspace*{0mm}
\end{figure}

\begin{figure}[h]
\vspace*{0mm}
\begin{minipage}[t]{.24\linewidth}
\subfloat[]{
\includegraphics[width=.9\textwidth]{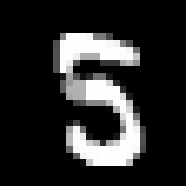}
\label{it14}}
\end{minipage}
\hfill
\begin{minipage}[t]{.24\linewidth}
\subfloat[]{
\includegraphics[width=.9\textwidth]{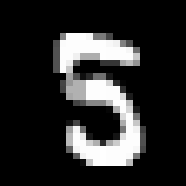}
\label{it15}}
\end{minipage}
\hfill
\begin{minipage}[t]{.24\linewidth}
\subfloat[]{
\includegraphics[width=.9\textwidth]{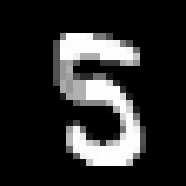}
\label{it10}}
\end{minipage}
\hfill
\begin{minipage}[t]{.24\linewidth}
\subfloat[]{
\includegraphics[width=.9\textwidth]{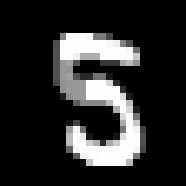}
\label{it11}}
\end{minipage}
\vspace*{0mm}
\caption{Adversarial samples applied to the DNN. (a) $14^{th}$ iteration of case I. (b) $15^{th}$ iteration of case I. (c) $10^{th}$ iteration of case II. (d) $11^{th}$ iteration of case II. }
\label{fig:dnnnpixel}
\vspace*{0mm}
\end{figure}

The baseline DNN classifies our test sample as a ``five'' with its associated probability equal to $98.79\%$, as shown in the blue-colored bars of Figure~\ref{fig:dnnnoise_both}. The selected target class is ``three'' for both the cases. The classification results of their respective adversarial images are shown in Figure~\ref{fig:dnnnoise_both} for both the cases. From the results in Table~\ref{tab:my_results}, we can observe that, having a small search window leads to obtaining a more robust attack, as compared to larger search windows. The generated adversarial examples are shown in \Cref{fig:dnnnpixel}.

\begin{table}[h]
\vspace*{2mm}
    \caption{Results of our simulations for the DNN. \\ \textbf{(Case I)} After 14 iterations, the probability of the target class has overcome the one of the initial class. Figure~\ref{it14} shows the sample at this stage (denoted as \textit{intermediate} in Figure~\ref{fig:dnnnoise_both}a). In the following iteration, the gap between the two classes increases, thus increasing the robustness, but also increasing the distance. The sample at this point (denoted as \textit{final} in Figure~\ref{fig:dnnnoise_both}a) corresponds to the attack output, since at the iteration 16 the distance falls above the threshold. \\ \textbf{(Case II)} After 11 iterations (denoted as \textit{final} in Figure~\ref{fig:dnnnoise_both}b), the sample (in Figure~\ref{it11}) is classified as a ``three''. Since at the iteration 12 the distance is already higher than $D_{MAX}$, Figure~\ref{it10} shows the sample at the 10$^{th}$, whose output probabilities are denoted as \textit{intermediate} in Figure~\ref{fig:dnnnoise_both}b. }
    \label{tab:my_results}
    \centering
    \resizebox{\linewidth}{!}{
\begin{tabular}{|c|c|c|c|c|c|c|}
\hline
      CASE &         ITER & \multicolumn{ 2}{|c|}{P MAX CLASS} & \multicolumn{ 2}{|c|}{P TARGET CLASS} &   DISTANCE \\
\hline
         I &          0 & \multicolumn{ 2}{|c|}{\textbf{98.79}} & \multicolumn{ 2}{|c|}{\textbf{0.89}} &          0 \\
\hline
         I &         14 & \multicolumn{ 2}{|c|}{\textbf{44.16}} & \multicolumn{ 2}{|c|}{\textbf{55.74}} &      20.18 \\
\hline
         I &         15 & \multicolumn{ 2}{|c|}{\textbf{36.25}} & \multicolumn{ 2}{|c|}{\textbf{63.67}} &      21.77 \\
\hline
         II &          0 & \multicolumn{ 2}{|c|}{\textbf{98.79}} & \multicolumn{ 2}{|c|}{\textbf{0.89}} &          0 \\
\hline
         II &         10 & \multicolumn{ 2}{|c|}{\textbf{57.53}} & \multicolumn{ 2}{|c|}{\textbf{42.01}} &      16.29 \\
\hline
         II &         11 & \multicolumn{ 2}{|c|}{\textbf{49.45}} & \multicolumn{ 2}{|c|}{\textbf{50.32}} &      21.19 \\
\hline
\end{tabular}  
}
    \vspace*{10pt}
\end{table}

\subsection{SDBN Under Attack}
\begin{figure}[h]
\centering
\vspace*{0mm}
\includegraphics[scale=0.55]{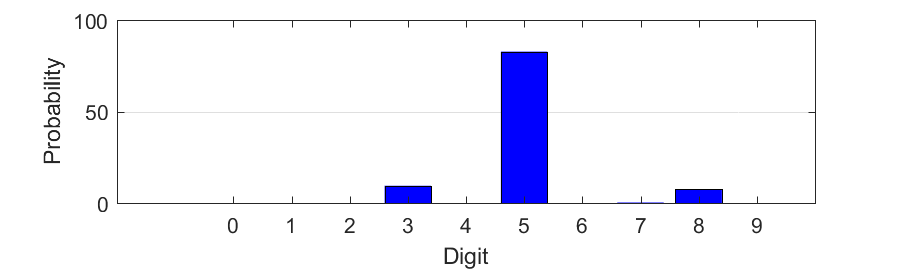}
\vspace*{-1mm}
\caption{Output probabilities of the SDBN for the original sample.}
\label{fig:histogram1}
\vspace*{0mm}
\end{figure}

Our baseline SDBN, without attack, classifies our test sample as a ``five'' with a probability equal to $82.69\%$. The complete set of initial ``clean-case'' output probabilities is shown in Figure~\ref{fig:histogram1}. We select the ``three'' as the target class.

\vspace*{6pt}

The results in Table~\ref{tab:results_SDBN} show that, in contrast to the attack applied to the DNN, for the case I:

\vspace*{3pt}

\begin{itemize}
    \item The SDBN output probabilities do not change monotonically when increasing the iterations of our algorithm.
    \item At the 20$^{th}$ iteration, the SDBN classifies the target class with a probability of $31.08\%$, while $D(X^*,X)=7.79$.
    \item At the other iterations, before and after iteration 20, the output probability of classifying the image as the original class still dominates.
\end{itemize}  

\noindent

\vspace*{6pt}

Meanwhile, for the case II, we observe that:

\vspace*{3pt}

\begin{itemize}
    \item At the 9$^{th}$ iteration, the SDBN misclassifies the image. The probability of classifying the image as a ``three'' is $50.60\%$, with a distance $D(X^*,X)=10.91$. As a side note, the probability of classifying the image as an ``eight'' is $49.40\%$.
    \item At the other iterations, before and after the iteration 7, the output probability of classifying the image as a ``five'' is higher than $50\%$.
\end{itemize}  

\begin{table}[h]
\vspace*{2mm}
    \caption{Results of our simulations for the SDBN.}
    \label{tab:results_SDBN}
    \vspace*{0mm}
    \centering
    \resizebox{\linewidth}{!}{
\begin{tabular}{|c|c|c|c|c|c|c|}
\hline
      CASE &         ITER & \multicolumn{ 2}{|c|}{P MAX CLASS} & \multicolumn{ 2}{|c|}{P TARGET CLASS} &   DISTANCE \\
\hline
         I &          0 & \multicolumn{ 2}{|c|}{\textbf{82.69}} & \multicolumn{ 2}{|c|}{\textbf{7.64}} &          0 \\
\hline
         I &         20 & \multicolumn{ 2}{|c|}{\textbf{60.29}} & \multicolumn{ 2}{|c|}{\textbf{31.08}} &      7.79 \\
\hline
         I &         21 & \multicolumn{ 2}{|c|}{\textbf{66.21}} & \multicolumn{ 2}{|c|}{\textbf{11.80}} &      8.15 \\
\hline
         II &          0 & \multicolumn{ 2}{|c|}{\textbf{82.69}} & \multicolumn{ 2}{|c|}{\textbf{7.64}} &          0 \\
\hline
         II &         9 & \multicolumn{ 2}{|c|}{\textbf{0}} & \multicolumn{ 2}{|c|}{\textbf{50.60}} &      10.91 \\
\hline
         II &         10 & \multicolumn{ 2}{|c|}{\textbf{64.94}} & \multicolumn{ 2}{|c|}{\textbf{12.03}} &      11.76 \\
\hline
\end{tabular}  
}
\end{table}

\vspace*{6pt}

\subsection{Comparative Discussion between SDBN and DNN}
We can observe how the DNN is vulnerable to the attacks generated by our algorithm, while the SDBN shows a very different response to the attack. The output probabilities of the SDBN do not follow the expected trend, but may sporadically lead to a misclassification if other conditions are satisfied as well. Each pixel of the image is converted to a spike train, thus a slight modification of the pixel intensity can have unexpected consequences, like a wrong feature detection. The SNN sensitivity of the targeted attack is clearly different from the DNN sensitivity for the similar case. Such a difference of robustness should be studied more carefully in future works.


\vspace*{6pt}

\section{Conclusions}
 
In this work, we studied the security vulnerabilities of SNNs, and compared them to DNNs under our attack methodology. However, there is still a long road for research to follow for analyzing and building robust/secure SNNs. Towards the conclusion of this work, we raise several new research questions like: ``What is hidden inside the SNNs that makes them more robust to targeted attacks, as compared to DNNs?'' ``Can certain specific properties of human brain's functionality be leveraged to build robust and self-healing machine learning algorithms?'' An extensive in-depth study of SNNs w.r.t. different security threats is crucial before adopting SNNs in safety-critical applications.


\vspace*{6pt}

\section*{Acknowledgments}

This work has been partially supported by the Doctoral College Resilient Embedded Systems which is run jointly by TU Wien's Faculty of Informatics and FH-Technikum Wien.

\vspace*{6pt}

\begin{refsize}
\bibliographystyle{abbrvnat}
\bibliography{main.bib}

\begin{thebibliography}{34}
\providecommand{\natexlab}[1]{#1}
\providecommand{\url}[1]{\texttt{#1}}
\expandafter\ifx\csname urlstyle\endcsname\relax
  \providecommand{\doi}[1]{doi: #1}\else
  \providecommand{\doi}{doi: \begingroup \urlstyle{rm}\Url}\fi

\bibitem[{Bagheri} et~al.(2018){Bagheri}, {Simeone}, and
  {Rajendran}]{Bagheri2018AdvTrainingSNN}
A.~{Bagheri}, O.~{Simeone}, and B.~{Rajendran}.
\newblock Adversarial training for probabilistic spiking neural networks.
\newblock In \emph{SPAWC}, 2018.

\bibitem[Bengio et~al.(2007)]{Bengio2007GreedyTraining}
Y.~Bengio et~al.
\newblock Greedy layer-wise training of deep networks.
\newblock In \emph{NIPS}. 2007.

\bibitem[{Davies} et~al.(2018)]{Davies2018Loihi}
M.~{Davies} et~al.
\newblock Loihi: A neuromorphic manycore processor with on-chip learning.
\newblock \emph{IEEE Micro}, 38\penalty0 (1):\penalty0 82--99, 2018.

\bibitem[Fatahi et~al.(2016)]{Fatahi2016evtMNIST}
M.~Fatahi et~al.
\newblock evt{\_}mnist: {A} spike based version of traditional {MNIST}.
\newblock \emph{CoRR}, abs/1604.06751, 2016.

\bibitem[Gerstner and Kistler(2002)]{Gerstner2002SpikingNeuronModel}
W.~Gerstner and W.~Kistler.
\newblock \emph{Spiking Neuron Models: An Introduction}.
\newblock Cambridge University Press, 2002.

\bibitem[Goh et~al.(2010)Goh, Thome, and Cord]{Goh2010BiasingRBM}
H.~Goh, N.~Thome, and M.~Cord.
\newblock {Biasing Restricted Boltzmann Machines to Manipulate Latent
  Selectivity and Sparsity}.
\newblock In \emph{{NIPS 2010 Workshop on Deep Learning and Unsupervised
  Feature Learning}}, 2010.

\bibitem[Goodfellow et~al.(2015)Goodfellow, Shlens, and
  Szegedy]{Goodfellow2015AdversarialExamples}
I.~Goodfellow, J.~Shlens, and C.~Szegedy.
\newblock Explaining and harnessing adversarial examples.
\newblock In \emph{ICLR}, 2015.

\bibitem[Heiberg et~al.(2013)Heiberg, Kriener, Tetzlaff, Einevoll, and
  Plesser]{Heiberg2013FiringrateSNN}
T.~Heiberg, B.~Kriener, T.~Tetzlaff, G.~T. Einevoll, and H.~E. Plesser.
\newblock Firing-rate models for neurons with a broad repertoire of spiking
  behaviors.
\newblock \emph{BMC Neuroscience}, 14:\penalty0 P317 -- P317, 2013.

\bibitem[Hinton and Salakhutdinov(2006)]{Hinton2006ReducingDimensionalityDNN}
G.~Hinton and R.~Salakhutdinov.
\newblock Reducing the dimensionality of data with neural networks.
\newblock \emph{Science (New York, N.Y.)}, pages 504--7, 2006.

\bibitem[Hinton et~al.(2006)Hinton, Osindero, and
  Teh]{Hinton2006FastLearningDBN}
G.~E. Hinton, S.~Osindero, and Y.-W. Teh.
\newblock A fast learning algorithm for deep belief nets.
\newblock \emph{Neural Computation}, 2006.

\bibitem[{Holmstrom} and {Koistinen}(1992)]{Holmstrom1992AdditiveNoiseBackprop}
L.~{Holmstrom} and P.~{Koistinen}.
\newblock Using additive noise in back-propagation training.
\newblock \emph{IEEE Transactions on Neural Networks}, 3\penalty0 (1):\penalty0
  24--38, 1992.

\bibitem[Kurakin et~al.(2017)Kurakin, Goodfellow, and
  Bengio]{Kurakin2017AdversarialPhysicalworld}
A.~Kurakin, I.~J. Goodfellow, and S.~Bengio.
\newblock Adversarial examples in the physical world.
\newblock In \emph{ICLR Workshop Track Proceedings}, 2017.

\bibitem[{Lisitsa} and {Zhilenkov}(2017)]{Lisitsa2017ProspectsSNN}
D.~{Lisitsa} and A.~A. {Zhilenkov}.
\newblock Prospects for the development and application of spiking neural
  networks.
\newblock In \emph{EIConRus}, pages 926--929, 2017.

\bibitem[Luo et~al.(2018)Luo, Liu, Wei, and Xu]{Luo2018ImperceptibleRobust}
B.~Luo, Y.~Liu, L.~Wei, and Q.~Xu.
\newblock Towards imperceptible and robust adversarial example attacks against
  neural networks.
\newblock In \emph{AAAI}, 2018.

\bibitem[Maas(1997)]{Maas1997ThirdGenerationSNN}
W.~Maas.
\newblock Networks of spiking neurons: The third generation of neural network
  models.
\newblock \emph{Trans. Soc. Comput. Simul. Int.}, 1997.

\bibitem[Madry et~al.(2018)Madry, Makelov, Schmidt, Tsipras, and
  Vladu]{Madry2018DLResistantAdversarial}
A.~Madry, A.~Makelov, L.~Schmidt, D.~Tsipras, and A.~Vladu.
\newblock Towards deep learning models resistant to adversarial attacks.
\newblock In \emph{ICLR}, 2018.

\bibitem[{Marchisio} et~al.(2019){Marchisio}, {Hanif}, {Khalid}, {Plastiras},
  {Kyrkou}, {Theocharides}, and {Shafique}]{Marchisio2019DL4EC}
A.~{Marchisio}, M.~A. {Hanif}, F.~{Khalid}, G.~{Plastiras}, C.~{Kyrkou},
  T.~{Theocharides}, and M.~{Shafique}.
\newblock Deep learning for edge computing: Current trends, cross-layer
  optimizations, and open research challenges.
\newblock In \emph{2019 IEEE Computer Society Annual Symposium on VLSI
  (ISVLSI)}, pages 553--559, 2019.

\bibitem[Marchisio et~al.(2019{\natexlab{a}})Marchisio, Nanfa, Khalid, Hanif,
  Martina, and Shafique]{Marchisio2019CapsAttacks}
A.~Marchisio, G.~Nanfa, F.~Khalid, M.~A. Hanif, M.~Martina, and M.~Shafique.
\newblock Capsattacks: Robust and imperceptible adversarial attacks on capsule
  networks.
\newblock \emph{ArXiv}, abs/1901.09878, 2019{\natexlab{a}}.

\bibitem[Marchisio et~al.(2019{\natexlab{b}})Marchisio, Nanfa, Khalid, Hanif,
  Martina, and Shafique]{Marchisio2019SNNUnderAttack}
A.~Marchisio, G.~Nanfa, F.~Khalid, M.~A. Hanif, M.~Martina, and M.~Shafique.
\newblock Snn under attack: are spiking deep belief networks vulnerable to
  adversarial examples?
\newblock \emph{ArXiv}, abs/1902.01147, 2019{\natexlab{b}}.

\bibitem[Merino et~al.(2018)Merino, Castrillejo, Pin, and
  Prats]{Merino2018WeightedContrastiveDivergence}
E.~R. Merino, F.~M. Castrillejo, J.~D. Pin, and D.~B. Prats.
\newblock Weighted contrastive divergence.
\newblock \emph{CoRR}, abs/1801.02567, 2018.

\bibitem[Merolla et~al.(2014)]{Merolla2014TrueNorth}
P.~A. Merolla et~al.
\newblock A million spiking-neuron integrated circuit with a scalable
  communication network and interface.
\newblock \emph{Science}, 2014.

\bibitem[M{\o}ller(1993)]{Mller1993ScaledConjugateGradient}
M.~F. M{\o}ller.
\newblock A scaled conjugate gradient algorithm for fast supervised learning.
\newblock \emph{Neural Networks}, 6:\penalty0 525--533, 1993.

\bibitem[O'Connor et~al.(2013)O'Connor, Neil, Liu, Delbruck, and
  Pfeiffer]{OConnor2013RTSNN}
P.~O'Connor, D.~Neil, S.-C. Liu, T.~Delbruck, and M.~Pfeiffer.
\newblock Real-time classification and sensor fusion with a spiking deep belief
  network.
\newblock \emph{Frontiers in Neuroscience}, 7:\penalty0 178, 2013.

\bibitem[Papernot et~al.(2017)Papernot, McDaniel, Goodfellow, Jha, Celik, and
  Swami]{Papernot2017PracticalBlackBox}
N.~Papernot, P.~McDaniel, I.~Goodfellow, S.~Jha, Z.~B. Celik, and A.~Swami.
\newblock Practical black-box attacks against machine learning.
\newblock In \emph{ASIA CCS}, 2017.

\bibitem[Ponulak and Kasinski(2011)]{Ponulak2011IntroductionSNN}
F.~Ponulak and A.~J. Kasinski.
\newblock Introduction to spiking neural networks: Information processing,
  learning and applications.
\newblock \emph{Acta neurobiologiae experimentalis}, 71 4:\penalty0 409--33,
  2011.

\bibitem[Shafahi et~al.(2019)]{Shafahi2018AdversarialInevitable}
A.~Shafahi et~al.
\newblock Are adversarial examples inevitable?
\newblock In \emph{ICLR}, 2019.

\bibitem[{Shafique} et~al.(2020){Shafique}, {Naseer}, {Theocharides}, {Kyrkou},
  {Mutlu}, {Orosa}, and {Choi}]{Shafique2020RobustML}
M.~{Shafique}, M.~{Naseer}, T.~{Theocharides}, C.~{Kyrkou}, O.~{Mutlu},
  L.~{Orosa}, and J.~{Choi}.
\newblock Robust machine learning systems: Challenges,current trends,
  perspectives, and the road ahead.
\newblock \emph{IEEE Design Test}, 37\penalty0 (2):\penalty0 30--57, 2020.

\bibitem[Siegert(1951)]{Siegert1951FirstPassageProbability}
A.~J.~F. Siegert.
\newblock On the first passage time probability problem.
\newblock \emph{Phys. Rev.}, 81:\penalty0 617--623, 1951.

\bibitem[Szegedy et~al.(2014)]{Szegedy2014IntriguingPropNN}
C.~Szegedy et~al.
\newblock Intriguing properties of neural networks.
\newblock In \emph{ICLR}, 2014.

\bibitem[Tavanaei et~al.(2019)Tavanaei, Ghodrati, Kheradpisheh, Masquelier, and
  Maida]{TAVANAEI2019DLSNN}
A.~Tavanaei, M.~Ghodrati, S.~R. Kheradpisheh, T.~Masquelier, and A.~Maida.
\newblock Deep learning in spiking neural networks.
\newblock \emph{Neural Networks}, 2019.

\bibitem[Tieleman and Hinton(2009)]{Tieleman2009FastWeights}
T.~Tieleman and G.~Hinton.
\newblock Using fast weights to improve persistent contrastive divergence.
\newblock In \emph{ICML}, 2009.

\bibitem[Vreeken(2003)]{Vreeken2003SNNIntroduction}
J.~Vreeken.
\newblock Spiking neural networks, an introduction.
\newblock Technical report, 2003.

\bibitem[Zhang and Jiang(2018)]{Zhang2018AdversarialOpportunitiesChallenges}
J.~Zhang and X.~Jiang.
\newblock Adversarial examples: Opportunities and challenges.
\newblock \emph{CoRR}, abs/1809.04790, 2018.

\bibitem[Zhang et~al.(2019)Zhang, Liu, Khalid, Hanif, Rehman, Theocharides,
  Artussi, Shafique, and Garg]{Zhang2019RobustML}
J.~J. Zhang, K.~Liu, F.~Khalid, M.~A. Hanif, S.~Rehman, T.~Theocharides,
  A.~Artussi, M.~Shafique, and S.~Garg.
\newblock Building robust machine learning systems: Current progress, research
  challenges, and opportunities.
\newblock \emph{Proceedings of the 56th Annual Design Automation Conference
  2019}, 2019.

\end{thebibliography}
\end{refsize}

\end{document}